*Original Article*

# Discretized Linear Regression and Multiclass Support Vector Based Air Pollution Forecasting Technique

M. Dhanalakshmi[1], V. Radha[2]

[1,2]Department of Computer Science, Avinashilingam Institute of Home Science and Higher Education for Women, Coimbatore.

[1]Corresponding Author : dhanalaxmi2289@gmail.com

*Abstract - Air pollution is a vital issue emerging from the uncontrolled utilization of traditional energy sources as far as developing countries are concerned. Hence, ingenious air pollution forecasting methods are indispensable to minimize the risk. To that end, this paper proposes an Internet of Things (IoT) enabled system for monitoring and controlling air pollution in the cloud computing environment. A method called Linear Regression and Multiclass Support Vector (LR-MSV) IoT-based Air Pollution Forecast is proposed to monitor the air quality data and the air quality index measurement to pave the way for controlling effectively. Extensive experiments carried out on the air quality data in the India dataset have revealed the outstanding performance of the proposed LR-MSV method when benchmarked with well-established state-of-the-art methods. The results obtained by the LR-MSV method witness a significant increase in air pollution forecasting accuracy by reducing the air pollution forecasting time and error rate compared with the results produced by the other state-of-the-art methods.*

## 1. Introduction

World health organization (WHO) disclosed that air pollution is highly susceptive to the sky-scraping environmental hazard to health and has resulted in a high mortality rate. Future research developments require machine learning techniques to predict air pollution quality monitoring and control. In addition to air pollution detection, the accuracy of quality being monitored has to be concentrated. This objective can be arrived at by examining the environment via IoT and reshaping neural networks.

## 2. Methodology

Nowadays, air pollution monitoring and control are used for numerous tasks measuring wind speed and direction, monitoring vehicle emissions etc. Therefore air pollution monitoring and control have to be performed in the earlier stage to control the hazard caused to humans, forests and animals. This air pollution monitoring and control issue is obviously addressed by crafting mechanisms by capturing the air quality data from various sensors (i.e., internet of things), storing and processing via a Cloud Computing environment. Air pollution monitoring and control via attention-based scheme and bidirectional RNN have been proposed in the recent literature

Despite satisfactory results provided by both classes of methods, both time and frequency aspects subdue noisy features and, owing to complicated data, compromise both accuracy and error rate. To address the above-said issues, in this work, a Linear Regression and Multiclass Support Vector (LR-MSV) IoT-based Air Pollution Forecast method for air pollution monitoring and control is proposed. Figure 1 shows the block diagram of the LR-MSV methodology.

As shown the above figure1, the LR-MSV method includes three stages. First, pre-processing of the air quality data in India to obtain noise-reduced pre-processed data is made computationally efficient by employing Wavelet Sliding Window-based Multi-resolute Pre-processing model.

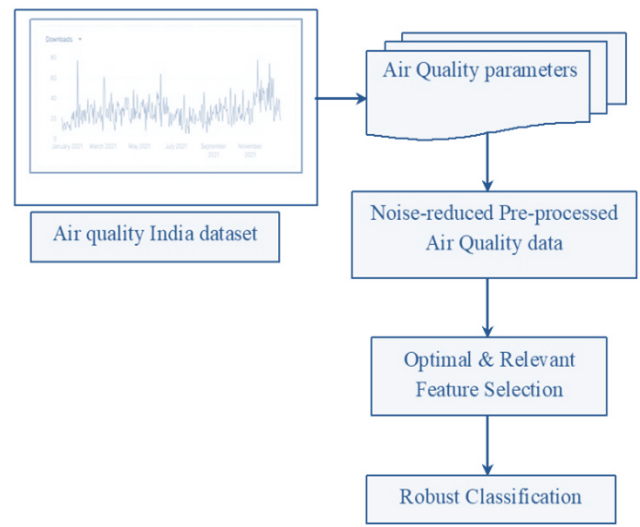

**Fig. 1 Block diagram of LR-MSV method**

Second, with the noise-reduced pre-processed air quality data, optimal and relevant features for further processing are made using Linear Regression and a Correlation-based Feature Selection model

Finally, with the identified relevant, a machine learning technique is applied to monitor air quality data through Air Quality Index efficiently and accordingly place the results in the AQI_bucket via multiclass support vectors. A detailed description of the proposed LR-MSV method is presented in the forthcoming sections, following a description of the Air Pollution Monitoring and Control System model.

## 2.1. Wavelet Sliding Window-based Multi-resolute Pre-Processing Model

Air pollution (AP) influences not only the environment but the body parts of humans and the respiratory system. Hence, systematic Air Quality (AQ) monitoring and control is required to evaluate the AP level and anticipate the pollutant concentrations with minimal noise. The wavelet decomposition model is an extensively utilized signal processing method in time series prediction. Its fundamental postulate remains in decomposing a non-smooth discrete time series air quality data into a blend of progressions with numerous high-frequency feature components 'H' and low-frequency coarse-grained component 'L'.

Here, the frequency of high-frequency feature components depends on the number of layers of the wavelet decomposition.

In this work, the Wavelet Sliding Window-based Multi-resolute Pre-processing model is applied to the raw air quality data obtained from the respective sensors to address the aspects involving both time and frequency. The multi-resolute here refers to the time and frequency while performing pre-processing via sliding window towards noise removal. The sliding window establishes per sample for each time instance 'T' and the samples of '$T_n$' utilizes the values as '[$T_{n-WS}, T_n$]', where '$T_n$' records the air quality values and window size 'WS' respectively. Figure 2 shows the structure of the Wavelet Sliding Window-based Multi-resolute Pre-processing model.

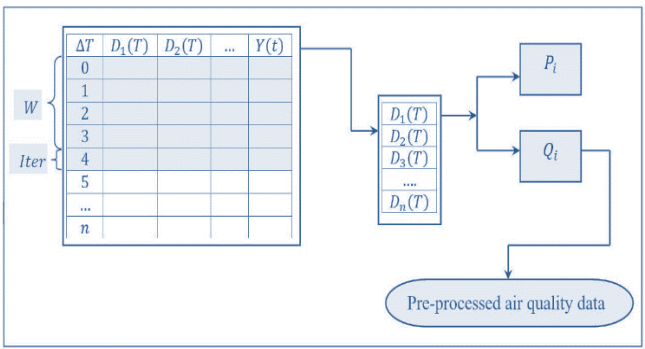

**Fig. 2 Wavelet Sliding Window-based Multi-resolute Pre-processing model**

As shown from the above figurative representation, the sliding window size value influences the number of time series samples and features present in the dataset. Owing to this, a small sliding window size refers to more air quality samples, whereas a larger sliding window size refers to fewer samples and more features. This sliding window model is then initially formulated as given below.

$$y(W + Iter) = f\left(T_0, \dots, T_W, D_{i,0}, \dots, D_{i,W}, T_{W+Iter}, Y_{W+Iter}\right) \quad (1)$$

$$y(W + Iter) = f\left(T_0, \dots, T_W, D_{i,0}, \dots, D_{i,W}, Y_0, \dots, Y_W, T_{W+Iter}, Y_{W+Iter}\right) \quad (2)$$

In the sliding window, wavelet decomposition is utilized by the cloud server upon recording each air quality data from the corresponding sensors (i.e., raw air quality time series data) that discretize high-frequency signal with high-frequency feature components from the low-frequency coarse-grained components to acquire more data features. The decomposition process for each feature is mathematically formulated as given below.

$$P_{i+1} = L(P_i) \quad (3)$$

$$Q_i = H(Q_i) \quad (4)$$

From the above two equations (3) and (4), '$P_i$', '$Q_i$' denotes the low-frequency coarse-grained component and high-frequency feature component, respectively, with '$L$' forming the low pass filter and '$H$' representing the high pass filter. During wavelet transform based pre-processing, each layer of the decomposed signal present in the sliding window is the bisection of the pre-decomposed air quality signal data. Hence, dual interpolation reconstructions are required to retrieve the signal length (i.e., pre-processed air quality data) and the reconstruction formula performed by the cloud server for each sensor is mathematically formulated as given below.

$$P_i = (L_2)^i P_i \quad (5)$$

$$Q_i = (L_2)^{i-1} H_2 Q_i \quad (6)$$

From the above two equations (5) and (6), '$L_2$' and '$H_2$' denotes the dual operators with which the cloud server distinguishes clearly between noise and significant information for air quality data at an hourly and daily level of various stations across multiple cities in India. The pseudocode representation is given below.

As given in the above Time and Frequency-based Sliding Window Pre-processing algorithm, the objective remains in returning the pre-processed air quality data by the cloud server for each corresponding sensor with minimum error; with this objective, initially sliding window for the respective sensors (i.e., device extracting air quality data like date, PM2.5, PM10, NO and so on) are modelled. Second, wavelet decomposition is formulated for the respective sensors according to low-frequency and high-frequency feature components. Third, with the dual operators, the cloud server returns the noise-reduced pre-processed air quality data for further processing.

| |
|---|
| **Input**: Dataset '$DS$', Cloud Server '$CS$', IoT Devices or Sensors '$S = S_1, S_2, ..., S_n$', Features '$F = F_1, F_2, ..., F_n$', Air Quality data '$D = D_1, D_2, ..., D_n$' |
| **Output**: Noise reduced pre-processed air quality data '$PD$' |
| Step 1: **Initialize** time instance '$T$' <br> Step 2: **Begin** <br> Step 3: **For** each Dataset '$DS$' (Air Quality data '$D = D_1, D_2, ..., D_n$') with Cloud Server '$CS$' and IoT Devices or Sensors '$S = S_1, S_2, ..., S_n$' <br> Step 4: Formulate a sliding window as in equations (1) and (2) <br> Step 5: **For** each Features '$F$' <br> Step 6: Perform decomposition as in equations (3) and (4) <br> Step 7: Model dual interpolation reconstructions to retrieve pre-processed air quality data as in equations (5) and (6) <br> Step 8: **Return** pre-processed air quality data '$PD$' <br> Step 9: **End for** <br> Step 10: **End for** <br> Step 11: **End** |

### 2.2. Linear Regression and Correlation-based Feature Selection

With the pre-processed air quality data, the next step remains to select the optimal features. Prevailing feature selection models like Monte Carlo [2] take no notice of the dependencies between the pre-processed features. A new model named Linear Regression and Correlation-based Feature Selection has been proposed based on the correlation between the features. Figure 4 shows the flow diagram of the Linear Regression and Correlation-based Feature Selection model.

As shown in the figure, let us assume that the independent pre-processed air quality data is '$PD = (PD_1, PD_2, ..., PD_n)$' with regression coefficients are '$\beta = (\beta_1, \beta_2, ..., \beta_n)$', then, the Linear Regression and Correlation-based Feature Selection prediction model is formulated as given below.

$$y_i = \beta_0 + \sum_{j=1}^{m} \beta_j + PD_j^i \quad (7)$$

From the above equation (7), '$PD_j^i$' forms the independent pre-processed air quality data, and '$\beta$' represents the regression coefficient and expected value of response (i.e., optimal and relevant feature selection) with the predictors (i.e., pre-processed air quality data) based on linear regression. '$y_i$' respectively. The optimum and relevant features are selected to minimize the sum of mean squared loss. This is mathematically formulated as given below.

$$\beta = argmin\, L(Dis, \beta) = argmin\, \sum_{i=1}^{n} (\beta \cdot PD_i - y_i)^2 \,(8)$$

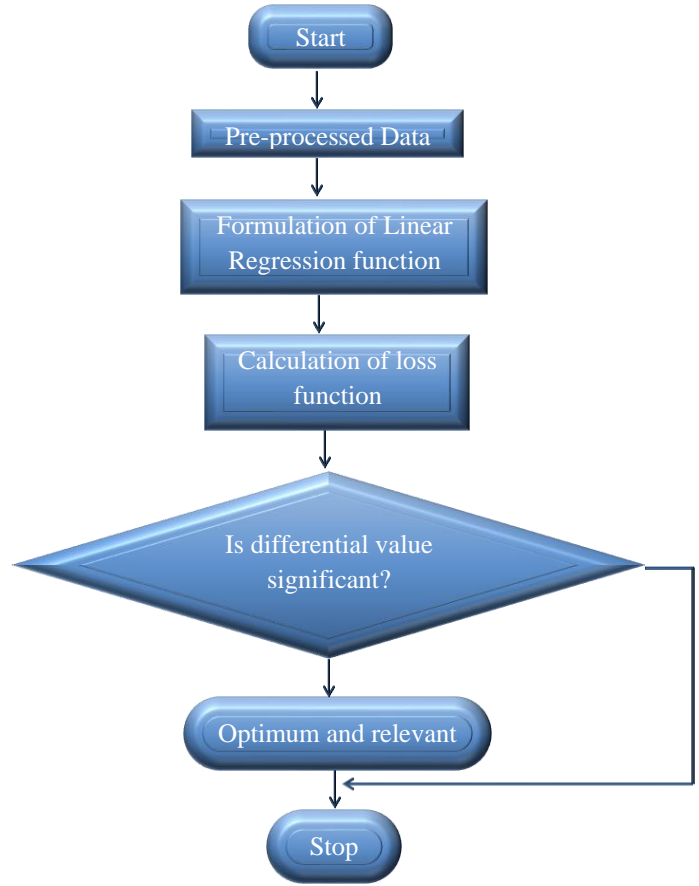

**Fig. 3 Flow of Linear Regression and Correlation-based Feature Selection**

From the above equation (8), the sum of mean squared loss '$L$' is estimated based on the distance '$D$' between two pre-processed data '$PD$' via regression coefficients '$\beta$' taken for selecting relevant feature. Finally, the optimum relevant features are selected using a gradient as given below.

$$\frac{\partial L(Dis, \beta)}{\partial \beta} = \frac{\partial (Y^T Y - Y^T PD\beta - \beta^T PD^T PD\beta)}{\partial \beta} \quad (9)$$

From the above equation (9), the optimum relevant features with minimum loss are obtained by the partial differentiating correlation between pre-processed data or features based on distance '$\partial L(Dis, \beta)$' for each regression coefficient '$\partial \beta$'. The pseudocode representation of Gradient Linear Regression-based Feature Selection is given below.

As given in the bel algorithm, Gradient Linear Regression-based Feature Selection is designed to improve the accuracy and time involved in air quality monitoring.

### 2.3 Multiclass Support Vector IoT-based Air Pollution Forecast model

Finally, in this section, with the relevant features selected (i.e., PM2.5, PM10, SO2, NOx, NH3, CO, O3), the classification of air quality data for the respective relevant features is made. The classification here is made through Air Quality Index (AQI) provided in table 1 to monitor the same and take preventive measures for controlling the same. Our work's classification uses a Multiclass Support Vector IoT-based Air Pollution Forecast model. Figure 4, given below, shows the structure of the Multiclass Support Vector IoT-based Air Pollution Forecast model.

**Algorithm.2 Gradient Linear Regression-based Feature Selection**

| |
|---|
| **Input**: Dataset '$DS$', Cloud Server '$CS$', IoT Devices or Sensors '$S = S_1, S_2, …, S_n$', Features '$F = F_1, F_2, …, F_n$', Air Quality data '$D = D_1, D_2, …, D_n$' |
| **Output**: Optimal and relevant feature selection |
| Step 1: **Initialize** pre-processed air quality data '$PD$', regression coefficients '$\beta = (\beta_1, \beta_2, …, \beta_n)$'<br>Step 2: Begin<br>Step 3: **For** each pre-processed air quality data '$PD$' with Cloud Server '$CS$' and IoT Devices or Sensors '$S = S_1, S_2, …, S_n$'<br>Step 4: Formulate linear regression function as given in equation (7)<br>Step 5: Evaluate the sum of mean squared loss as given in equation (8)<br>Step 6: Obtain optimum relevant features using the gradient function as given in equation (9)<br>Step 7: **Return** relevant features ($RF$)<br>Step 8: **End for**<br>Step 9: **End** |

As shown in the above figure, given a dataset '$DS$' with relevant features selected '$RF = \{RF_1, RF_2, …, RF_n\}$', '$RF_i \in R^n$' belong to '$C$' different classes (i.e., '$C = 6$'), forming a tuple. '$(RF_i, Y_i)$'. In addition, let '$Y_i \in \{0, 0 − 0.25, 0.25 − 0.50, 0.50 − 0.75, 0.75 − 1, > 1\}$' representing 6 different classes, namely, good, satisfactory, moderate, poor, very poor and severe, respectively, is considered as its class labels. The recommendation confinements of the linear classifier are then mathematically formulated as given below.

$$W^T RF + B \qquad (10)$$

From the above equation (10), '$W$' denotes the weight vector and '$B$' forms the bias vector, respectively. Several linear separators are present. However, the SVM design objective remains in determining a decision boundary that is maximally far away from any data point. This distance from the Decision boundary (i.e., AQI) to the closest data point (i.e., RF) determines the margin of the classifier. The air quality index for each sampled data is measured as given below.

$$AQI = Avg(PM2.5, PM10, SO2, NOx, NH3) + Max(CO, O3) \qquad (11)$$

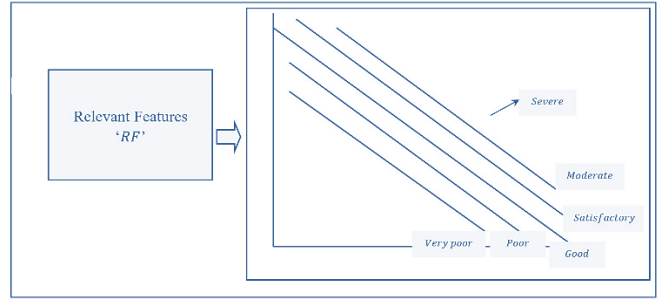

**Fig. 4 Multiclass Support Vector IoT-based Air Pollution Forecast model**

**Algorithm. 3 Multiclass Support Vector IoT-based Air Pollution Forecast**

| |
|---|
| **Input**: Dataset '$DS$', Cloud Server '$CS$', IoT Devices or Sensors '$S = S_1, S_2, …, S_n$', Features '$F = F_1, F_2, …, F_n$', Air Quality data '$D = D_1, D_2, …, D_n$' |
| **Output**: Robust classification |
| Step 1: **Initialize** relevant features ($RF$)<br>Step 2: **Begin**<br>Step 3: **For** each relevant features '$RF$' with Cloud Server '$CS$' and IoT Devices or Sensors '$S = S_1, S_2, …, S_n$'<br>Step 4: Formulate recommendation confinements of the linear classifier as in equation (10)<br>Step 5: Measure Air Quality Index as in equation (11)<br>Step 6: **If** '$W^T RF_i(AQI) + b = 0$'<br>Step 7: **Then** AQI_Bucket '$AQI_B \to Good$'<br>Step 8: **End if**<br>Step 9: **If** '$W^T RF_i + b(AQI) > 0$ and $W^T RF_i + b(AQI) < 0.50$'<br>Step 10: **Then** AQI_Bucket '$AQI_B \to Satisfactory$'<br>Step 11: **End if**<br>Step 12: **If** '$W^T RF_i + b(AQI) > 0.50$ and $W^T RF_i + b(AQI) < 1$'<br>Step 13: **Then** AQI_Bucket '$AQI_B \to Moderate$'<br>Step 14: **End if**<br>Step 15: **If** '$W^T RF_i + b(AQI) > 0$ and $W^T RF_i + b(AQI) < −0.50$'<br>Step 16: **Then** AQI_Bucket '$AQI_B \to Poor$'<br>Step 17: **End if**<br>Step 18: **If** '$W^T RF_i + b(AQI) > −0.50$ and $W^T RF_i + b(AQI) < −1$'<br>Step 19: **Then** AQI_Bucket '$AQI_B \to Very\ Poor$'<br>Step 20: **End if**<br>Step 21: **If** '$W^T RF_i + b(AQI) > 1$'<br>Step 22: **Then** AQI_Bucket '$AQI_B \to Severe$'<br>Step 23: **End if**<br>Step 24: **End for**<br>Step 25: **End** |

From the above equation (11), the air quality index value '$AQI$' is measured based on the average values of PM2.5, PM10, SO2, NOx, NH3 and the maximum values of CO and O3, respectively. These points are referred to as the support vectors. Finally, the resultant of AQI_Bucket is estimated to measure the air quality so that control measures can be made according to the pollutants. The pseudocode representation of the Multiclass Support Vector IoT-based Air Pollution Forecast is given below.

As given in the above Multiclass Support Vector IoT-based Air Pollution Forecast algorithm, the objective is to forecast air pollution with minimum error.

## 3. Experimental Setup

In this section, the performance of air pollution monitoring and control using IoT in a cloud computing environment method called Linear Regression and Multiclass Support Vector (LR-MSV) is performed using the Java interfaces and CloudSim simulator. To measure the LR-MSV method, air quality data in India via https://www.kaggle.com/rohanrao/air-quality-data-in-india is used. First, dataset details are provided. Following this, experiments were conducted on factors such as air pollution forecasting time, air pollution forecasting accuracy and error rate concerning different air quality samples. A fair comparison is made with the existing methods, Integrated Multiple Directed Attention and Variational Auto Encoder (VAE) (IMD-VAE) [1] and Bidirectional Recurrent Neural Network [bidirectional RNN] [2] for a simulation of 10 runs.

### 3.1. Dataset Details

The Air Quality in India dataset contains air quality data and AQI (Air Quality Index) measured on both hourly and daily basis of several stations across multiple cities in India

**Table 1 Air Quality in India dataset description**

| S. No | Features | Description |
|---|---|---|
| 1 | City | City name |
| 2 | Date | Date of occurrence |
| 3 | PM 2.5 | Particulate Matter 2.5 |
| 4 | PM 10 | Particulate Matter 10 |
| 5 | NO | Nitric Oxide |
| 6 | NO2 | Nitric dioxide |
| 7 | NOx | Any nitric x-oxide |
| 8 | NH3 | Ammonia |
| 9 | CO | Carbon monoxide |
| 10 | SO2 | Sulphur dioxide |
| 11 | O3 | Ozone |
| 12 | C6H6 | Benzene |
| 13 | C7H8 | Toluene |
| 14 | C8H10 | Xylene |

### 3.2. Performance Analysis of Air Pollution Forecasting Time

Air pollution forecasting refers to applying science and technology to predict air pollution composition in the atmosphere for any location taken into consideration along with its respective time. Mainstream pollution forecasting methods tend to utilize air quality index to indicate pollution levels, and also, in our work, it has been employed using a multiclass support vector. The air pollution forecasting time is mathematically formulated as given below.

$$APF_{time} = \sum_{i=1}^{n} D_i * Time\,[W^T RF + B] \qquad (12)$$

From the above equation (12), the air pollution forecasting time '$APF_{time}$' is measured based on the air quality sample data involved in the simulation process. '$D_i *$' and the time consumed in the forecasting process via multiclass support vectors ' $Time\,[W^T RF + B]$ '. It is measured in terms of milliseconds (ms). Table 2 below shows the performance analysis of the proposed air pollution monitoring and controlling method, LR-MSV, for air pollution forecasting time. The proposed method is applied to the air quality data obtained from different sensors and stored in the cloud server. In this work, the simulation results of the proposed air pollution monitoring and controlling method are compared with other state-of-the-art methods, IMD-VAE [1] and bidirectional RNN [2], which used similar air quality data in the India dataset.

**Table 2. Tabulation for air pollution forecasting time**

| Air quality data | Air pollution forecasting time (ms) | | |
|---|---|---|---|
| | LR-MSV | IMD-VAE | bidirectional RNN |
| 2000 | 960 | 1100 | 1300 |
| 4000 | 1025 | 1255 | 1635 |
| 6000 | 1085 | 1315 | 1855 |
| 8000 | 1235 | 1535 | 1925 |
| 10000 | 1415 | 1895 | 2235 |
| 12000 | 1635 | 2135 | 2455 |
| 14000 | 1915 | 2325 | 2835 |
| 16000 | 2245 | 2635 | 3015 |
| 18000 | 2585 | 3025 | 3455 |
| 20000 | 3025 | 3355 | 3825 |

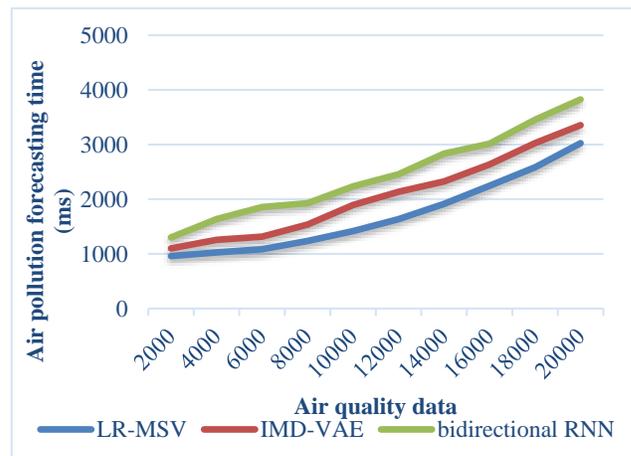

**Fig. 5 Graphical representation of air pollution forecasting time**

Figure 5 above shows the performance metric of air pollution forecasting time for varying air quality data obtained from distinct sensors, each obtaining different air quality data in the range of 2000 to 20000 collected at different time intervals. From the figure, it is inferred that the air pollution forecasting time increases with the increase in the number of air quality data. This is because increasing the number of air quality data causes an increase in the number of data involved in pre-processing, which also increases the air pollution forecasting time. However, With '2000' air quality data involved in simulation and the time consumed in air pollution forecasting for single data being '0.48$ms$' using LR-MSV, the overall air pollution forecasting time was observed to be '960$ms$', time consumed in air pollution forecasting being '0.55$ms$' using [1] and '0.65$ms$' using [2], the overall air pollution forecasting time was observed to be '1100$ms$' and '1300$ms$'. From the result, the air pollution forecasting time via pre-processing was found to be minimum using LR-MSV when compared to [1] and [2]. The reason behind the improvement was due to the application of the Time and Frequency-based Sliding Window Pre-processing algorithm.

### 3.3. Performance Analysis of Air Pollution Forecasting Accuracy

The second parameter of significance for air pollution monitoring and control is air pollution forecasting accuracy. This parameter is highly significant because the more accurate the air pollution forecasting being made earlier the control measure done and therefore preventing hazardous disease. The mathematical formulation for air pollution forecasting accuracy is given below.

$$APF_{acc} = \sum_{i=1}^{n} \frac{D_{FAcc}}{D_i} * 100 \qquad (13)$$

**Table 3. Tabulation for air pollution forecasting accuracy**

| Air quality data | Air pollution forecasting accuracy (%) | | |
|---|---|---|---|
| | LR-MSV | IMD-VAE | bidirectional RNN |
| **2000** | 92.25 | 85.75 | 83.75 |
| **4000** | 91.45 | 83.15 | 81.45 |
| **6000** | 91 | 81 | 78.15 |
| **8000** | 90.25 | 78.35 | 75.35 |
| **10000** | 88.35 | 78 | 71 |
| **12000** | 85.25 | 77.35 | 70.25 |
| **14000** | 83.15 | 75.25 | 68.35 |
| **16000** | 83 | 75 | 65 |
| **18000** | 82.55 | 73.15 | 63.15 |
| **20000** | 80 | 73 | 60 |

From the above equation (13), the air pollution forecasting accuracy. '$APF_{acc}$' is measured based on the air quality sample data considered for simulation. '$D_i$' and the air quality data was forecasted accurately. '$D_{FAcc}$'. It is measured in terms of percentage (%). Table 3 compares the proposed air pollution monitoring and control system with state of arts IMD-VAE [1] and bidirectional RNN [2]. The traditional air pollution monitoring and control methods utilized air quality data, and the results are tabulated in Table 3. From Table 3, it is very clear that the air pollution monitoring system achieves maximum accuracy.

Second, figure 6 given above demonstrates the air pollution forecasting accuracy for 20000 varying numbers of air quality data. From the figure, it is inferred that the accuracy of air pollution forecasting decreases with the increase in air quality data. This is because a small portion of the presence of noise or artefacts is said to be retained during the pre-processing stage, and accuracy variation is also said to be identified. However, simulations conducted with '2000' air quality data '1845' air quality data were correctly detected as it is using LR-MSV, whereas '1715' and '1675' were correctly detected using [1] and [2]. From this analysis, the air pollution forecasting accuracy was found to be '92.25%', '85.75%' and '83.75%' using LR-MSV, [1] and [2] respectively. From this result, it is inferred that the air pollution forecasting accuracy using the LR-MSV method is said to be comparatively better than [1] and [2]. The reason behind the improvement is due to the incorporation of the Gradient Linear Regression-based Feature Selection algorithm. By applying this algorithm, air quality data forecasted accurately that form the basis for early and precise air pollution monitoring and control is said to be obtained via the Gradient Linear Regression function for each pre-processed air quality data. Then, the sum of the mean squared loss for each feature with the adjacent is made for selecting the best feature for air pollution monitoring. With this, the air pollution forecasting accuracy using the LR-MSV method is said to be improved by 11% compared to [1] and 22% compared to [2].

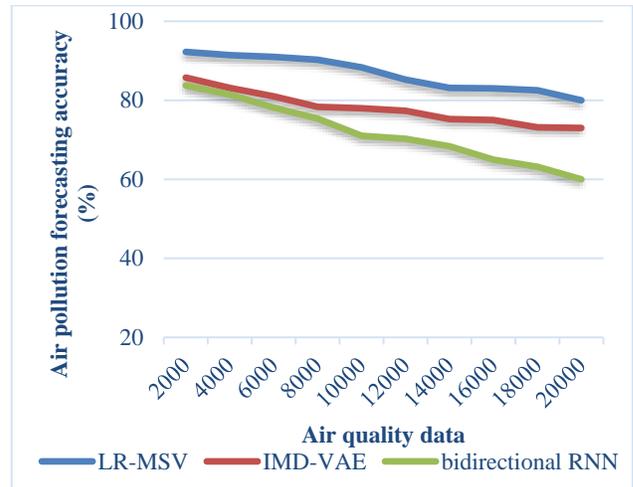

**Fig. 6 Graphical representation of air pollution forecasting accuracy**

### 3.4. Performance Analysis of Error Rate

Finally, the metric of significance that has to be measured for air pollution monitoring and control is the error rate. The error rate is mathematically formulated as given below.

$$ER = \sum_{i=1}^{n} \frac{D_{FWrongly}}{D_i} * 100 \qquad (14)$$

From the above equation (14), the error rate '$ER$' is measured based on the air quality data samples involved in the simulation process. '$D_i$' and the air quality data was forecasted wrongly. '$D_{FWrongly}$'. This is measured in terms of percentage (%). Table 4 compares the proposed air pollution monitoring and control error rate with other state-of-the-art methods [1] and [2], respectively.

**Table 4. Tabulation for error rate**

| Air quality data | Error rate (%) | | |
|---|---|---|---|
| | LR-MSV | IMD-VAE | bidirectional RNN |
| **2000** | 4.25 | 6.25 | 7.75 |
| **4000** | 4.75 | 7.15 | 9.25 |
| **6000** | 5.15 | 8.35 | 9.85 |
| **8000** | 5.35 | 8.85 | 10.35 |
| **10000** | 5.85 | 9.25 | 10.85 |
| **12000** | 6.05 | 9.55 | 11.25 |
| **14000** | 6.55 | 9.85 | 11.85 |
| **16000** | 7.25 | 10.35 | 12.45 |
| **18000** | 8.35 | 10.85 | 12.85 |
| **20000** | 9.25 | 11 | 13 |

Finally, figure 7 given above shows the error rate for 20000 distinct air quality data obtained at different time instances. With error rate increased with the increasing air quality samples but was found to be comparatively reduced using LR-MSV upon comparison with [1] and [2]. However, simulations performed with 20000 samples observed that 85 samples were wrongly forecasted using LR-MSV, 125 samples were wrongly forecasted using [1], and 155 samples were wrongly forecasted using [2]. With this, the overall error rate was 4.25% using the LR-MSV method, 6.25% using [1] and 7.75% using [2]. The minimum error rate contribution using the LR-MSV method was due to the application of the Multiclass Support Vector IoT-based Air Pollution Forecast algorithm. By applying this algorithm, the air quality index was first measured and then supported vectors were obtained based on multiclass, therefore reducing the error rate using the

LR-MSV method by 32% compared to [1] and 43% compared to [2], respectively.

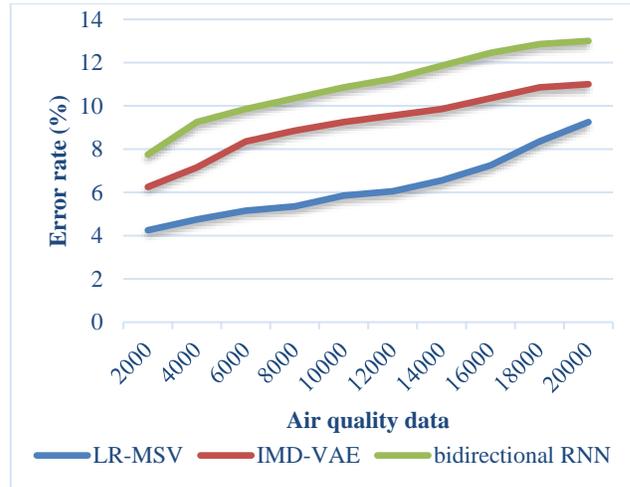

**Fig. 7 Graphical representation of error rate**

## 4. Conclusion

The machine learning techniques serve as a refining service to confront air pollution forecast and control, progressively increasing the attention of researchers and academics. Inferences from previous research work specify that there is a requirement to design an effective method to provide accurate forecasting and control measures to be taken accordingly based on the air pollution via air quality index. Specifically, there is a requirement to address time and accuracy involving air pollution forecasting to prevent hazardous effects on humans. Hence, this work aims to address air pollution monitoring and control via machine learning technique to develop a Linear Regression and Multiclass Support Vector (LR-MSV) IoT-based Air Pollution Forecast method that provides forecasts in an accurate and timely manner with a minimum error rate. The experimental results show that the LR-MSV method can get better results in terms of air pollution forecasting accuracy, time and error rate on aid quality data in the India dataset than others, which fully shows that applying it to machine learning can improve the forecasting accuracy and therefore paving the way for significant control.